\documentclass[10pt,twocolumn,letterpaper]{article}

\usepackage[pagenumbers]{iccv} 

\usepackage{multirow}
\usepackage{multicol}
\usepackage{adjustbox}
\usepackage{algorithm}
\usepackage{algorithmic}
\usepackage{colortbl}
\usepackage[accsupp]{axessibility}
\usepackage[dvipsnames*,svgnames]{xcolor}
\usepackage{colortbl}

\definecolor{iccvblue}{rgb}{0.21,0.49,0.74}
\usepackage[pagebackref,breaklinks,colorlinks,allcolors=iccvblue]{hyperref}

\newcommand{\ccb}{\cellcolor{blue!10}}

\def\paperID{11} 
\def\confName{ICCV}
\def\confYear{2025}

\title{SciFlow: \underline{S}emantic \underline{C}ross \underline{I}nterference for
\\Self-Supervised Optical Flow Domain Generalization}

\author{
Jamie Menjay Lin$^1$~~~
Jisoo Jeong$^2$~~~
Hong Cai$^2$~~~
Kai Wang$^1$~~~
Fatih Porikli$^2$~~~
\smallskip
\\
$^1$Qualcomm Technologies, Inc. ~~~
$^2$Qualcomm AI Research$^{\dagger}$~~~
\\
\smallskip
{\tt\small\{jmlin, jisojeon, hongcai, kwang, fporikli\}@qti.qualcomm.com \vspace{-12pt}}
}

\begin{document}
\maketitle
\begin{abstract}

Motions of objects and scenes carry essential intelligence in video understanding, offering rich cues for interpreting dynamic settings and interactions. Due to the cost and scarcity of high-quality annotation or ground truth of pixel-wise optical flow, however, motion estimation models are typically trained in synthetic domains while deployed in real-world domains. Addressing synthetic-to-real domain generalization challenges has been crucial for developing practical solutions in diverse open-world use cases. 

This paper introduces SciFlow, a simple yet effective, network-agnostic, training-based approach that leverages self-supervised learning to generalize motion estimation across synthetic and open-world domains. Specifically, SciFlow imposes semantic interference from open-world images onto synthetic images during training, blending in-domain features with cross-domain interference, which enables the network to adapt to the real-world domains. Additionally, SciFlow utilizes geometric consistency to ensure validity of the self-supervision. Our experiment results show that SciFlow not only significantly enhances model robustness amidst domain variations, but also remarkably enables synthetic-to-real domain generalization without requiring any ground truth in the open world.

\end{abstract}

\begin{figure}[]
\begin{center}$
\centering
\begin{tabular}{cccc}
\hspace{-0.3cm} \includegraphics[width=1.8cm,height=1.8cm]{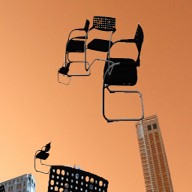}
&\hspace{-0.45cm} \includegraphics[width=1.8cm,height=1.8cm]{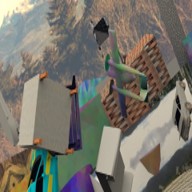} &
\multicolumn{2}{c}{\hspace{-0.35cm}\includegraphics[width=3.7cm,height=1.8cm]{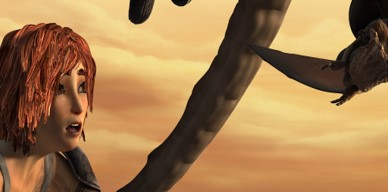}} \vspace{-0.1cm}\\
 \hspace{-0.3cm}  (a) & \hspace{-0.45cm} (b) & \multicolumn{2}{c}{\hspace{-0.35cm}  (c)}\\
\multicolumn{2}{c}{\hspace{-0.3cm}\includegraphics[width=3.7cm,height=1.8cm]{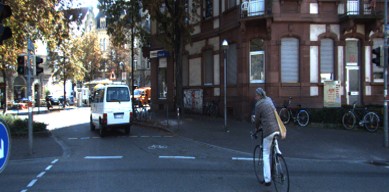}} &
\multicolumn{2}{c}{\hspace{-0.4cm}\includegraphics[width=3.7cm,height=1.8cm]{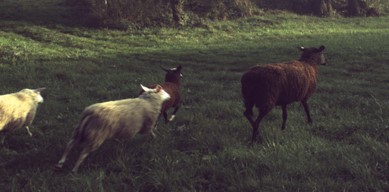}} \vspace{-0.1cm}\\
\multicolumn{2}{c}{\hspace{-0.3cm} (d)} & \multicolumn{2}{c}{\hspace{-0.4cm} (e)}
\end{tabular}$
\end{center}
\vspace{-17pt}
\caption{\small \textbf{Domain shifts in terms of motions, lighting, and focus (sharpness) across synthetic and open-world datasets, and across driving-specific and general-activity datasets.} (a)-(c) Samples in the synthetic FlyingChairs~\cite{dosovitskiy2015flownet}, FlyingThings3D~\cite{mayer2016large}, and Sintel~\cite{butler2012naturalistic} datasets that often present rich contrasts in color intensity and object motions. (d) A sample in the real-world KITTI~\cite{geiger2013vision, menze2015object, Menze2015ISA} dataset consisting of driving scenes of rigid (non-deformable) vehicle shapes, simple vehicle edges, and steady vehicle and global (ego) motions. (e) Sample in the open-world SlowFlow~\cite{Janai2017SlowFlow} dataset containing diversity in outdoor and indoor activities, and scenes under various image qualities and conditions (e.g., lighting and sharpness).}
\vspace{-4mm}
\label{fig:experiment_WSVDExamples}
\end{figure}

\begin{figure*}[t]
\centering
\includegraphics[width=0.8\linewidth]{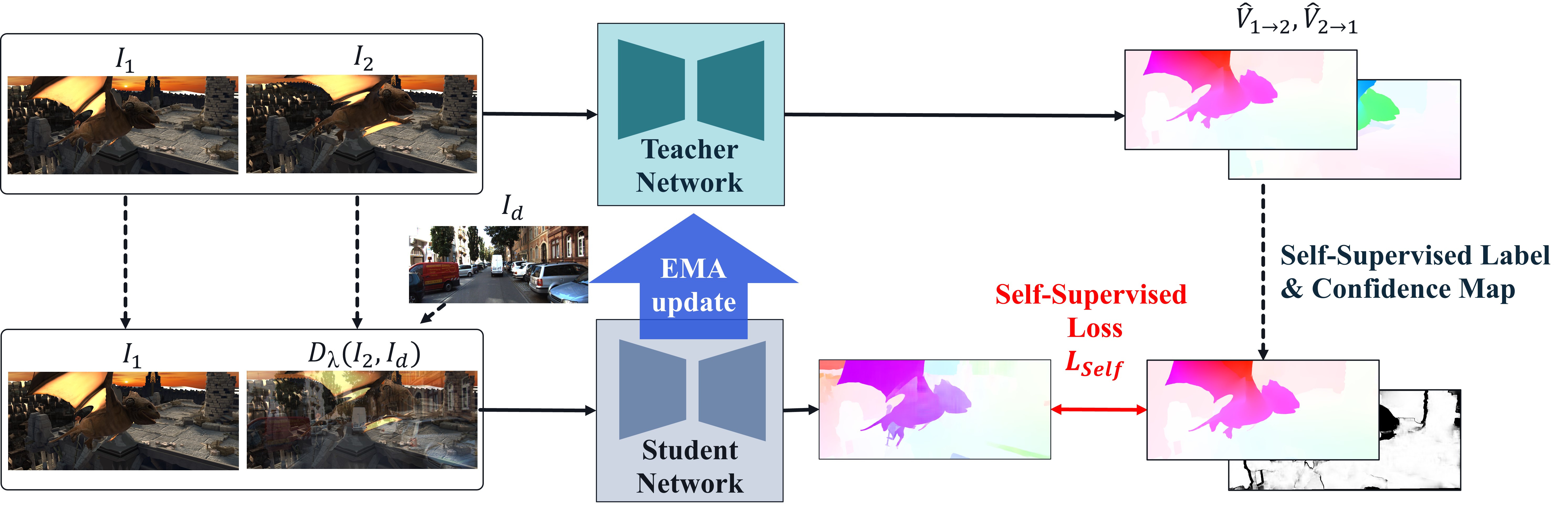}
\vspace{-7pt}
\caption{\small \textbf{Our proposed SciFlow for self-supervised domain generation.} SciFlow leverages semantic cross-domain interference and incorporates additional techniques, including Exponential Moving Average (EMA) and confidence masking, in a unified self-supervised framework. The teacher is initialized with the student weights and takes continuous EMA updates from the student during self-supervised learning. While the teacher takes (clean) samples in previously trained domains, the student takes samples with semantic cross-domain interference and its predictions are compared against the self-supervised labels predicted by the teacher to create the self-supervised loss.}
\vspace{-8pt}
\label{fig:method}
\end{figure*}    
\section{Introduction}
\label{sec:intro}

{\let\thefootnote\relax\footnotetext{{
\hspace{-6.5mm} $\dagger$ Qualcomm AI Research is an initiative of Qualcomm Technologies, Inc.}}}

Motion (optical flow) estimation plays a crucial role in video understanding and underpins numerous vision tasks, including object tracking, frame interpolation~\cite{jeong2024ocai}, and video compression~\cite{lu2019dvc}. 
Recent advancement has been driven by deep learning, encompassing convolutional networks~\cite{dosovitskiy2015flownet, ilg2017flownet, ranjan2017optical, sun2018pwc}. optimization-inspired models~\cite{teed2020raft}, transformers~\cite{huang2022flowformer}, and diffusion models~\cite{saxena2023surprising, luo2024flowdiffuser}. 
Despite such progress, acquiring dense motion ground truth for real-world scenes remains a challenge, limiting the effectiveness of supervised learning. Consequently, synthetic datasets~\cite{dosovitskiy2015flownet, mayer2016large, butler2012naturalistic} have become the primary training source for optical flow models.


Semi- and self-supervised learning methods~\cite{jeong2022imposing, jeong2023distractflow, han2022realflow, jeong2024ocai} have been actively explored to overcome the limitations of relying solely on labeled synthetic data for optical flow estimation. These approaches aim to leverage unlabeled target-domain data and small amounts of labeled real-world data to improve generalization. Among recent efforts, RAFT-OCTC~\cite{jeong2022imposing} introduces transformation consistency as a regularization strategy for semi-supervised training. DistractFlow~\cite{jeong2023distractflow} enforces consistency between flow predictions from original and strongly augmented image pairs to improve robustness. RealFlow~\cite{han2022realflow} and OCAI~\cite{jeong2024ocai} generate augmented, unlabeled image pairs to support semi-supervised learning pipelines. ADFactory~\cite{Ling_2024_CVPR} takes a different route by reconstructing static scenes via neural rendering and augmenting them with dynamic objects to synthesize optical flow data in open-world settings. While these methods have shown promise, they often struggle to generalize to unseen domains or require computationally expensive data augmentation and rendering pipelines.

In this paper, we propose SciFlow, a Semantic Cross-Domain Interference method for self-supervised optical flow generalization. SciFlow is a simple yet effective approach to synthetic-to-real domain adaptation that avoids the need for real or pseudo optical flow labels, as well as complex reconstruction or rendering pipelines.

During self-supervised training, SciFlow augments image pairs by compositing open-world images onto synthetic scenes, enabling the network to adapt to real-world statistics while mitigating domain shift. When combined with existing techniques, SciFlow improves performance on standard benchmarks like KITTI~\cite{geiger2013vision} and delivers robust, accurate flow estimation on diverse, in-the-wild samples.

\vspace{2pt} Our main contributions are as follows: 

\begin{itemize} \item We introduce SciFlow, a network-agnostic method that leverages semantic cross-domain interference in self-supervised learning to improve optical flow domain generalization. \item SciFlow achieves strong cross-domain performance without requiring real-world ground truth, addressing both domain shift and label scarcity in open-world scenarios. \end{itemize}

\begin{figure*}[ht]
\begin{center}$
\centering
\begin{tabular}{cccc}

\hspace{-0.3cm} \textbf{\scriptsize{Arbitrary Open-World Test Sample}} 
& \hspace{-0.3cm} \textbf{\scriptsize{Pre SciFlow (Baseline)}} 
& \hspace{-0.3cm} \textbf{\scriptsize{Post SciFlow (Ours)}} 
& \hspace{-0.3cm} \textbf{\scriptsize{Difference between Predictions}} \\
\hspace{-0.3cm} \includegraphics[width=4.2cm,height=2.2cm]{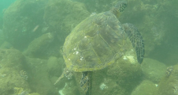}
&\hspace{-0.45cm} \includegraphics[width=4.2cm,height=2.2cm]{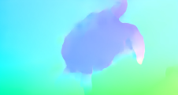} 
&\hspace{-0.45cm} \includegraphics[width=4.2cm,height=2.2cm]{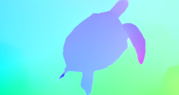} 
&\hspace{-0.45cm} \includegraphics[width=4.2cm,height=2.2cm]{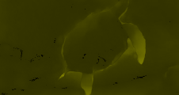} \vspace{-0.1cm} \\
\hspace{-0.3cm} \includegraphics[width=4.2cm,height=2.2cm]{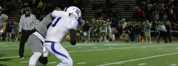}
&\hspace{-0.45cm} \includegraphics[width=4.2cm,height=2.2cm]{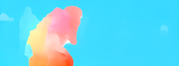} 
&\hspace{-0.45cm} \includegraphics[width=4.2cm,height=2.2cm]{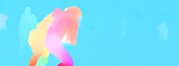} 
&\hspace{-0.45cm} \includegraphics[width=4.2cm,height=2.2cm]{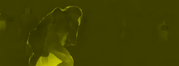} \vspace{-0.1cm} \\
\hspace{-0.3cm} \includegraphics[width=4.2cm,height=2.2cm]{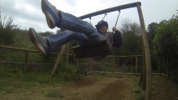}
&\hspace{-0.5cm} \includegraphics[width=4.2cm,height=2.2cm]{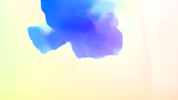} 
&\hspace{-0.45cm} \includegraphics[width=4.2cm,height=2.2cm]{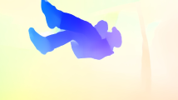} 
&\hspace{-0.45cm} \includegraphics[width=4.2cm,height=2.2cm]{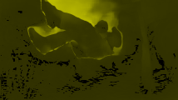} \vspace{-0.1cm} \\
\hspace{-0.3cm} \includegraphics[width=4.2cm,height=2.2cm]{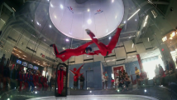}
&\hspace{-0.45cm} \includegraphics[width=4.2cm,height=2.2cm]{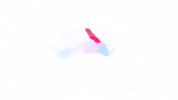} 
&\hspace{-0.45cm} \includegraphics[width=4.2cm,height=2.2cm]{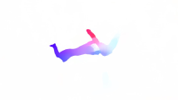} 
&\hspace{-0.45cm} \includegraphics[width=4.2cm,height=2.2cm]{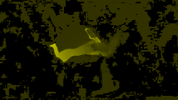} \vspace{-0.1cm} \\
\hspace{-0.3cm} \includegraphics[width=4.2cm,height=2.2cm]{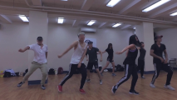}
&\hspace{-0.45cm} \includegraphics[width=4.2cm,height=2.2cm]{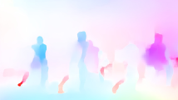} 
&\hspace{-0.45cm} \includegraphics[width=4.2cm,height=2.2cm]{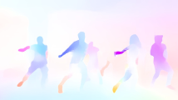} 
&\hspace{-0.45cm} \includegraphics[width=4.2cm,height=2.2cm]{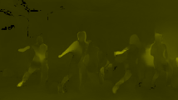} 
\vspace{-19pt}
\end{tabular}$
\end{center}
\caption{\small Qualitative results without and with SciFlow (in 2nd and 3rd columns, respectively) of generalizing MobileFlow from synthetic domain to unseen, open-world domain. Notably, SciFlow enables robust, accurate estimation on diverse samples in the wild.}
\label{fig:experiment_visual}
\end{figure*}


\vspace{18pt}
\section{Preliminaries}
\label{sec:pre}
\vspace{-3pt}





In this section, we introduce the in-domain distraction technique from DistractFlow~\cite{jeong2023distractflow} and draw inspiration from it to develop a new cross-domain semantic interference strategy in SciFlow.

\textbf{Notations:} Given two consecutive video frames, $I_{1}$ and $I_{2}$, we denote the (forward) optical flow as $V_{1 \rightarrow 2}$, which maps each pixel in $I_{1}$ to its corresponding location in $I_{2}$. The optical flow model is denoted as $f$.

Given a training image pair, DistractFlow~\cite{jeong2023distractflow} introduces a third image from the same domain or dataset and mixes it with one or both images in the original pair. This augmentation is used in a semi-supervised setting, where the optical flow prediction from the augmented pair is encouraged to match the prediction from the original pair. More formally, the self-supervision loss and total loss are defined as: 
\begin{equation}\label{eq:distractflow_lself}
\begin{split}
& L_{self} = ||[M_{conf} \geq  \tau](f(I_{1},I_{2})- f(I_{1}, D_{\lambda}(I_{2},I_{d})))||_{1}, \\
& L_{total} = L_{sup} + w_{self} \cdot L_{self},\\[-13pt]
\end{split}
\end{equation}
where ($I_{1}, D_{\lambda}(I_{2},I_{d})$) is the augmented image pair, $D_{\lambda}$ is the distraction function that mixes two images with ratio $\lambda$, $M_{conf}$ is a forward-backward consistency-based confidence mask against a threshold $\tau$, and $w_{self}$ is a weighting hyperparameter. See~\cite{jeong2023distractflow} for further details.

Training with such augmented samples improves model robustness to semantic variations, though only within the same domain.
\section{Method}
\label{sec:method}
\vspace{-3pt}
In this section, we present the method of SciFlow. We start with an overview of the proposed self-supervised learning strategy. Then we discuss the specific design in SciFlow that enables synthetic-to-real domain generalization.

\subsection{Overall Training Procedure}\vspace{-2pt}

Consider two dataset groups: a synthetic dataset $D_{GT}$ with ground truth labels, and a real-world dataset $D_{NGT}$ without ground truth.
We begin by applying supervised curriculum learning~\cite{DBLP:conf/icml/HacohenW19, sebe2022curriculum} on $D_{GT}$, starting with samples exhibiting simpler motion and scene complexity, and gradually progressing to more challenging examples. After leveraging $D_{GT}$ across varying complexity levels, we transition to the next training phase. In this second phase, the goal is to adapt the optical flow model to real-world domains. To this end, we introduce SciFlow, which performs self-supervised fine-tuning on the unlabeled real-world dataset $D_{NGT}$.

\subsection{Self-Supervised Learning with SciFlow}\vspace{-2pt}
In this self-supervised phase, we adopt a teacher-student learning framework, where both networks share the same architecture, denoted as $f_{T}$ (teacher) and $f_{S}$ (student). This setup has proven effective for training optical flow models~\cite{jeong2024ocai}.

Given two consecutive video frames $I_{1}$ and $I_{2}$ with ground truth from $D_{GT}$, and a third image $I_{d}$ from a \textit{different} domain in $D_{NGT}$ (without ground truth), we now construct two image pairs: the original pair ($I_{1}$, $I_{2}$), and the cross-domain pair ($I_{1}$, $D_{\lambda}(I_{2},I_{d})$). The latter is generated via semantic cross-domain interference, as our SciFlow paper title suggests, where $D_{\lambda}(I_{2},I_{d}) = \lambda \cdot I_{2} + (1-\lambda) \cdot I_{d}$ blends the second frame with $I_{d}$ using a mixing ratio $\lambda$ sampled from a beta distribution~\cite{yun2019cutmix, DBLP:conf/iclr/ZhangCDL18}.

The teacher $f_{T}$ processes the original pair, while the student $f_{S}$ is trained on the more challenging cross-interfered pair. To align the student’s prediction with the teacher’s, we define the self-supervision loss as:
\begin{equation} \label{sup}
\begin{split}
L_{self} = [M_{conf} \geq \tau]||f_{S}(I_{1}, D_{\lambda}(I_{2},I_{d})) - f_{T}(I_{1}, I_{2})||_{1}
\end{split}
\end{equation}
where the confidence mask $M_{conf}$ and threshold $\tau$ follow the definition in Section~\ref{sec:pre}. No ground truth from the target domain is required.

Unlike DistractFlow~\cite{jeong2023distractflow}, which uses only in-domain augmentation, our semantic cross-domain interference enables training with both original synthetic pairs from $D_{GT}$ and new pairs incorporating features from the target domain $D_{NGT}$. This helps the model adapt to real-world characteristics without introducing instability from abrupt domain shifts.

Finally, we update the student $f_{S}$ using the self-supervised loss $L_{self}$, and apply Exponential Moving Average (EMA) to update the teacher $f_{T}$, promoting stable and consistent training.

\section{Experiments}
\label{sec:exp}
\vspace{-3pt}

\begin{table}[t]
\begin{center}
\caption{\textbf{Quantitative evaluation of self-supervision methods on KITTI benchmarks.} More descriptions on used datasets can be found in Section \ref{exp_training}.
Training options of Exponential Moving Average (EMA), confidence-base loss masking, and semantic cross-domain interference are denoted as E, C, and I, and entries of the proposed SciFlow setting are highlighted in grey. Brackets [] indicate a small number (200) of KITTI train samples are included in the $D_{GT}$ dataset before SciFlow self supervision on the large WSVD dataset.
}
\vspace{-8pt}
\label{tab:sciflow}
\adjustbox{max width=0.48\textwidth}
{
\begin{tabular}{|c|c|c|c|c|c||cc|}
\hline
\multirow{2}{*}{Network (GMACs)} & \multirow{2}{*}{Method} & \multirow{2}{*}{$E$} & \multirow{2}{*}{$C$} & \multirow{2}{*}{$I$} & {Unlabeled} &   \multicolumn{2}{|c|}{KITTI 15} \\
& & & & & Dataset & (Fl-epe) & (Fl-all)\\ 
\hline
\hline
\multirow{6}{*}{\begin{tabular}[c]{@{}c@{}} RAFT \cite{teed2020raft} \\ {(808.9)} \end{tabular}} & Supervised & & & &   & 5.04 & 17.4 \\
\cline{2-8}
& DistractFlow & & & \checkmark & Sintel+KM &3.01 & 11.7 \\
\cline{2-8}
& \multirow{4}{*}{SciFlow (ours)} &  & & & KM  & \multicolumn{2}{c|}{Diverged} \\
&  & \checkmark  & & & KM   & 2.85 & 10.1 \\
& & \checkmark & \checkmark  & & KM  & 2.81 & 9.7 \\
& & \ccb \checkmark  &\ccb \checkmark  &\ccb \checkmark  &\ccb KM  &\ccb 2.27 &\ccb 8.4 \\
\hline
\multirow{5}{*}{\begin{tabular}[c]{@{}c@{}} MobileFlow \cite{Lin_2024_CVPR} \\{(78.8)} \end{tabular}} & Supervised & & & &   & 8.20 & 22.6 \\
\cline{2-8}
& \multirow{4}{*}{SciFlow (ours)} &  & & & KM  & \multicolumn{2}{c|}{Diverged} \\
& & \checkmark  & \checkmark & & KM   & 3.24 & 9.7 \\
& &\ccb \checkmark &\ccb \checkmark  &\ccb \checkmark &\ccb KM  &\ccb 2.78 &\ccb 8.95 \\
\cline{3-8}
& &\ccb  \checkmark  &\ccb  \checkmark  &\ccb  \checkmark  &\ccb  WSVD &\ccb  [1.65] &\ccb  [6.24] \\
\hline
\end{tabular}
}
\vspace{-4mm}
\end{center}
\end{table}

\subsection{Datasets, Networks, and Training Setups}\vspace{-3pt}
\label{exp_training}
\textbf{Datasets:} We use FlyingChairs~\cite{dosovitskiy2015flownet}, FlyingThings3D~\cite{mayer2016large}, Sintel~\cite{butler2012naturalistic}, HD1K~\cite{kondermann2016hci}, KITTI~\cite{geiger2013vision, menze2015object, Menze2015ISA}, TartanAir~\cite{tartanair2020iros}, and Spring~\cite{Mehl2023_Spring} as the $D_{GT}$ datasets for supervised learning. Besides, we use Web Stereo Video Dataset (WSVD)~\cite{wang2019web}, which contains 689 videos of totally 1.5 million frames, as our open-world $D_{GT}$ dataset for self-supervised learning without ground truth. We use DAVIS (2017) \cite{Pont-Tuset_arXiv_2017} and SlowFlow~\cite{Janai2017SlowFlow} for visual evaluation.

\textbf{Networks:} 
We use two distinct network architectures, RAFT~\cite{teed2020raft} as a full-capacity model and MobileFlow~\cite{Lin_2024_CVPR} as a light-weight mobile-friendly model. The former serves as a reference for quantitative model accuracy. The latter is our target model serving both quantitative and qualitative evaluation for domain generalization.

\textbf{Training \#1:}
We first train the models on FlyingChairs and FlyingThings3D in the supervised learning phase. Then we fine tune the model with SciFlow for domain generalization on Sintel and/or KITTI Multiview as the unlabeled datasets without using any ground truth or pseudo labels.

\textbf{Training \#2:}
For further synthetic-to-real domain generalization, besides $D_{GT}$ datasets including FlyingChairs, FlyingThings, TartanAir, Sintel, KITTI (multiview) and Spring, our $D_{NGT}$ set starts to include a larger WSVD dataset, which serves as the large open-world datasets to represent in-the-wild samples arbitrarily captured without ground truth. This experiment is intended for full generalization into the in-the-wild domains, as the WSVD dataset contains various scenes, including those in indoor and outdoor, simulating diverse activities in everyday mobile (imperfect) conditions.

\begin{figure}[t]
\centering
\includegraphics[width=0.95\linewidth]{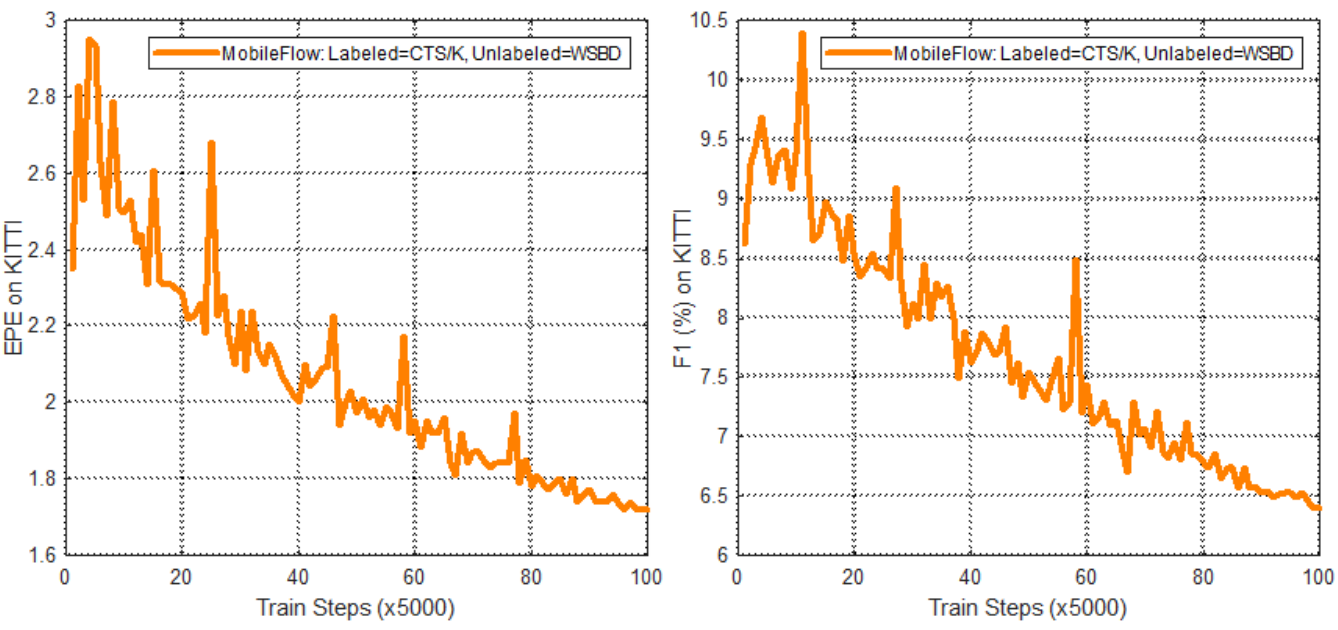}
\vspace{-3mm}
\caption{Effectiveness of SciFlow using WSVD as the real-world self-supervision dataset and using KITTI as the proxy dataset for validation.}
\vspace{-4mm}
\label{fig:experiment_TrainCurves}
\end{figure}

\subsection{Results}
\vspace{-1mm}

Table~\ref{tab:sciflow} summarizes our quantitative evaluation of the baseline and SciFlow on the KITTI benchmarks. 
Two network architectures, RAFT and MobileFlow, are shown in the top and bottom halves, respectively. The supervised-learning baselines for each architecture is provided for reference. 
First, our SciFlow method trained with EMA (E), Confident-based loss masking (C), and semantic cross-domain interference (I) achieves significant gains over corresponding baselines. 
Secondly, for the light-weight MobileFlow at much smaller model complexity, SciFlow enables even higher accuracy over the large open-world WSVD dataset without needing any ground truth during the self-supervised learning.

In Fig~\ref{fig:experiment_TrainCurves}, we demonstrate the effectiveness of domain generalization through validation on KITTI, which serves as the proxy dataset, when using SciFlow to finetune a lightweight MobileFlow \cite{Lin_2024_CVPR} model from synthetic (plus 200 samples in KITTI) to open-world WSVD \cite{wang2019web} domains. (Left) Validation EPE on the proxy KITTI dataset over training steps. (Right) Validation Fl-all error on the proxy KITTI dataset over training steps.

Fig~\ref{fig:experiment_visual} visualizes qualitative results over arbitrary samples from open-world datasets DAVIS \cite{Pont-Tuset_arXiv_2017} and SlowFlow~\cite{Janai2017SlowFlow} using the light-weight MobileFlow supervised on synthetic datasets and self-supervised on WSVD. The results demonstrate effectiveness of SciFlow, over the baseline, for domain generalization into unseen open-world domains under conditions such as low light, motion blur, and high noise.

\vspace{-2mm}
\section{Conclusion}
\label{sec:conclusion}
\vspace{-1mm}


SciFlow offers a simple yet effective solution to the challenge of synthetic-to-real domain generalization in motion estimation. By introducing semantic cross-domain interference during self-supervised learning, SciFlow enables models to effectively adapt to real-world scenarios without requiring any ground truth from the target domain. Its simplicity and architecture-agnostic nature make it broadly applicable across models and use cases, providing a practical path toward more robust and accurate motion estimation for video understanding in complex, dynamic environments.

\newpage

{
    \small
    \bibliographystyle{ieeenat_fullname}
    \bibliography{main}

@String(CVPR= {IEEE Conf. Comput. Vis. Pattern Recog.})

@String(ICLR = {Int. Conf. Learn. Represent.})

@String(CVPR  = {CVPR})

@String(ICLR  = {ICLR})

@inproceedings{jeong2023distractflow,
  title={Distractflow: Improving optical flow estimation via realistic distractions and pseudo-labeling},
  author={Jeong, Jisoo and Cai, Hong and Garrepalli, Risheek and Porikli, Fatih},
  booktitle={Proceedings of the IEEE/CVF Conference on Computer Vision and Pattern Recognition},
  pages={13691--13700},
  year={2023}
}

@inproceedings{dosovitskiy2015flownet,
  title={Flownet: Learning optical flow with convolutional networks},
  author={Dosovitskiy, Alexey and Fischer, Philipp and Ilg, Eddy and Hausser, Philip and Hazirbas, Caner and Golkov, Vladimir and Van Der Smagt, Patrick and Cremers, Daniel and Brox, Thomas},
  booktitle={Proceedings of the IEEE/CVF International Conference on Computer Vision},
  pages={2758--2766},
  year={2015}
}

@inproceedings{ilg2017flownet,
  title={Flownet 2.0: Evolution of optical flow estimation with deep networks},
  author={Ilg, Eddy and Mayer, Nikolaus and Saikia, Tonmoy and Keuper, Margret and Dosovitskiy, Alexey and Brox, Thomas},
  booktitle={Proceedings of the IEEE/CVF Conference on Computer Vision and Pattern Recognition},
  pages={2462--2470},
  year={2017}
}

@inproceedings{ranjan2017optical,
  title={Optical flow estimation using a spatial pyramid network},
  author={Ranjan, Anurag and Black, Michael J},
  booktitle={Proceedings of the IEEE/CVF Conference on Computer Vision and Pattern Recognition},
  pages={4161--4170},
  year={2017}
}

@inproceedings{sun2018pwc,
  title={Pwc-net: Cnns for optical flow using pyramid, warping, and cost volume},
  author={Sun, Deqing and Yang, Xiaodong and Liu, Ming-Yu and Kautz, Jan},
  booktitle={Proceedings of the IEEE/CVF Conference on Computer Vision and Pattern Recognition},
  pages={8934--8943},
  year={2018}
}

@inproceedings{teed2020raft,
  title={Raft: Recurrent all-pairs field transforms for optical flow},
  author={Teed, Zachary and Deng, Jia},
  booktitle={Proceedings of the European Conference on Computer Vision},
  pages={402--419},
  year={2020},
  organization={Springer}
}

@inproceedings{mayer2016large,
  title={A large dataset to train convolutional networks for disparity, optical flow, and scene flow estimation},
  author={Mayer, Nikolaus and Ilg, Eddy and Hausser, Philip and Fischer, Philipp and Cremers, Daniel and Dosovitskiy, Alexey and Brox, Thomas},
  booktitle={Proceedings of the IEEE/CVF Conference on Computer Vision and Pattern Recognition},
  pages={4040--4048},
  year={2016}
}

@inproceedings{butler2012naturalistic,
  title={A naturalistic open source movie for optical flow evaluation},
  author={Butler, Daniel J and Wulff, Jonas and Stanley, Garrett B and Black, Michael J},
  booktitle={Proceedings of the European Conference on Computer Vision},
  pages={611--625},
  year={2012},
  organization={Springer}
}

@article{geiger2013vision,
  title={Vision meets robotics: The kitti dataset},
  author={Geiger, Andreas and Lenz, Philip and Stiller, Christoph and Urtasun, Raquel},
  journal={The International Journal of Robotics Research},
  volume={32},
  number={11},
  pages={1231--1237},
  year={2013},
  publisher={Sage Publications Sage UK: London, England}
}

@inproceedings{menze2015object,
  title={Object scene flow for autonomous vehicles},
  author={Menze, Moritz and Geiger, Andreas},
  booktitle={Proceedings of the IEEE/CVF Conference on Computer Vision and Pattern Recognition},
  pages={3061--3070},
  year={2015}
}

@inproceedings{kondermann2016hci,
  title={The hci benchmark suite: Stereo and flow ground truth with uncertainties for urban autonomous driving},
  author={Kondermann, Daniel and Nair, Rahul and Honauer, Katrin and Krispin, Karsten and Andrulis, Jonas and Brock, Alexander and Gussefeld, Burkhard and Rahimimoghaddam, Mohsen and Hofmann, Sabine and Brenner, Claus and others},
  booktitle={Proceedings of the IEEE/CVF Conference on Computer Vision and Pattern Recognition Workshops},
  pages={19--28},
  year={2016}
}

@inproceedings{wang2019web,
  title={Web stereo video supervision for depth prediction from dynamic scenes},
  author={Wang, Chaoyang and Lucey, Simon and Perazzi, Federico and Wang, Oliver},
  booktitle={2019 International Conference on 3D Vision (3DV)},
  pages={348--357},
  year={2019},
  organization={IEEE}
}

@inproceedings{lu2019dvc,
  title={Dvc: An end-to-end deep video compression framework},
  author={Lu, Guo and Ouyang, Wanli and Xu, Dong and Zhang, Xiaoyun and Cai, Chunlei and Gao, Zhiyong},
  booktitle={Proceedings of the IEEE/CVF Conference on Computer Vision and Pattern Recognition},
  pages={11006--11015},
  year={2019}
}

@inproceedings{huang2022flowformer,
  title={FlowFormer: A Transformer Architecture for Optical Flow},
  author={Huang, Zhaoyang and Shi, Xiaoyu and Zhang, Chao and Wang, Qiang and Cheung, Ka Chun and Qin, Hongwei and Dai, Jifeng and Li, Hongsheng},
  booktitle={Proceedings of the European Conference on Computer Vision},
  year={2022}
}

@inproceedings{jeong2022imposing,
  title={Imposing Consistency for Optical Flow Estimation},
  author={Jeong, Jisoo and Lin, Jamie Menjay and Porikli, Fatih and Kwak, Nojun},
  booktitle={Proceedings of the IEEE/CVF Conference on Computer Vision and Pattern Recognition},
  pages={3181--3191},
  year={2022}
}

@inproceedings{yun2019cutmix,
  title={Cutmix: Regularization strategy to train strong classifiers with localizable features},
  author={Yun, Sangdoo and Han, Dongyoon and Oh, Seong Joon and Chun, Sanghyuk and Choe, Junsuk and Yoo, Youngjoon},
  booktitle={Proceedings of the IEEE/CVF International Conference on Computer Vision},
  pages={6023--6032},
  year={2019}
}

@INPROCEEDINGS{Janai2017SlowFlow,
  author = {Joel Janai and Fatma Güney and Jonas Wulff and Michael Black and Andreas Geiger},
  title = {Slow Flow: Exploiting High-Speed Cameras for Accurate and Diverse Optical Flow Reference Data},
  booktitle = {Proceedings of the IEEE/CVF Conference on Computer Vision and Pattern Recognition},
  year = {2017}
}

@article{han2022realflow,
  title={RealFlow: EM-based Realistic Optical Flow Dataset Generation from Videos},
  author={Han, Yunhui and Luo, Kunming and Luo, Ao and Liu, Jiangyu and Fan, Haoqiang and Luo, Guiming and Liu, Shuaicheng},
  journal={arXiv preprint arXiv:2207.11075},
  year={2022}
}

@inproceedings{jeong2024ocai,
  title={Ocai: Improving optical flow estimation by occlusion and consistency aware interpolation},
  author={Jeong, Jisoo and Cai, Hong and Garrepalli, Risheek and Lin, Jamie Menjay and Hayat, Munawar and Porikli, Fatih},
  booktitle={Proceedings of the IEEE/CVF Conference on Computer Vision and Pattern Recognition},
  pages={19352--19362},
  year={2024}
}

@InProceedings{Lin_2024_CVPR,
    author    = {Lin, Jamie Menjay and Jeong, Jisoo and Cai, Hong and Garrepalli, Risheek and Wang, Kai and Porikli, Fatih},
    title     = {SciFlow: Empowering Lightweight Optical Flow Models with Self-Cleaning Iterations},
    booktitle = {Proceedings of the IEEE/CVF Conference on Computer Vision and Pattern Recognition (CVPR) Workshops},
    month     = {June},
    year      = {2024},
    pages     = {2162-2171}
}

@inproceedings{Menze2015ISA,
  author = {Moritz Menze and Christian Heipke and Andreas Geiger},
  title = {Joint 3D Estimation of Vehicles and Scene Flow},
  booktitle = {ISPRS Workshop on Image Sequence Analysis (ISA)},
  year = {2015}
}

@article{tartanair2020iros,
  title =   {TartanAir: A Dataset to Push the Limits of Visual SLAM},
  author =  {Wang, Wenshan and Zhu, Delong and Wang, Xiangwei and Hu, Yaoyu and Qiu, Yuheng and Wang, Chen and Hu, Yafei and Kapoor, Ashish and Scherer, Sebastian},
  booktitle = {2020 IEEE/RSJ International Conference on Intelligent Robots and Systems (IROS)},
  year =    {2020}
}

@InProceedings{Mehl2023_Spring,
    author    = {Lukas Mehl and Jenny Schmalfuss and Azin Jahedi and Yaroslava Nalivayko and Andr\'es Bruhn},
    title     = {Spring: A High-Resolution High-Detail Dataset and Benchmark for Scene Flow, Optical Flow and Stereo},
    booktitle = {Proc. IEEE/CVF Conference on Computer Vision and Pattern Recognition (CVPR)},
    year      = {2023}
}

@article{saxena2023surprising,
  title={The surprising effectiveness of diffusion models for optical flow and monocular depth estimation},
  author={Saxena, Saurabh and Herrmann, Charles and Hur, Junhwa and Kar, Abhishek and Norouzi, Mohammad and Sun, Deqing and Fleet, David J},
  journal={Advances in Neural Information Processing Systems},
  volume={36},
  pages={39443--39469},
  year={2023}
}

@inproceedings{luo2024flowdiffuser,
  title={Flowdiffuser: Advancing optical flow estimation with diffusion models},
  author={Luo, Ao and Li, Xin and Yang, Fan and Liu, Jiangyu and Fan, Haoqiang and Liu, Shuaicheng},
  booktitle={Proceedings of the IEEE/CVF Conference on Computer Vision and Pattern Recognition},
  pages={19167--19176},
  year={2024}
}

@inproceedings{DBLP:conf/iclr/ZhangCDL18,
  author       = {Hongyi Zhang and
                  Moustapha Ciss{\'{e}} and
                  Yann N. Dauphin and
                  David Lopez{-}Paz},
  title        = {mixup: Beyond Empirical Risk Minimization},
  booktitle    = {6th International Conference on Learning Representations, {ICLR} 2018,
                  Vancouver, BC, Canada, April 30 - May 3, 2018, Conference Track Proceedings},
  publisher    = {OpenReview.net},
  year         = {2018},
  url          = {https://openreview.net/forum?id=r1Ddp1-Rb},
  bibsource    = {dblp computer science bibliography, https://dblp.org}
}

@inproceedings{DBLP:conf/icml/HacohenW19,
  author       = {Guy Hacohen and
                  Daphna Weinshall},
  editor       = {Kamalika Chaudhuri and
                  Ruslan Salakhutdinov},
  title        = {On The Power of Curriculum Learning in Training Deep Networks},
  booktitle    = {Proceedings of the 36th International Conference on Machine Learning,
                  {ICML} 2019, 9-15 June 2019, Long Beach, California, {USA}},
  series       = {Proceedings of Machine Learning Research},
  volume       = {97},
  pages        = {2535--2544},
  publisher    = {{PMLR}},
  year         = {2019},
  url          = {http://proceedings.mlr.press/v97/hacohen19a.html},
  bibsource    = {dblp computer science bibliography, https://dblp.org}
}

@article{sebe2022curriculum,
  title={Curriculum Learning: A Survey},
  author={Sebe, Nicoleta and Ionescu, Radu Tudor and Soviany, Petru},
  journal={International Journal of Computer Vision},
  year={2022},
  doi={10.1007/s11263-022-01611-x}
}

@InProceedings{Ling_2024_CVPR,
    author    = {Ling, Han and Sun, Quansen and Sun, Yinghui and Xu, Xian and Li, Xinfeng},
    title     = {ADFactory: An Effective Framework for Generalizing Optical Flow with NeRF},
    booktitle = {Proceedings of the IEEE/CVF Conference on Computer Vision and Pattern Recognition (CVPR)},
    month     = {June},
    year      = {2024},
    pages     = {20591-20600}
}

@article{Pont-Tuset_arXiv_2017,
  author = {Jordi Pont-Tuset and Federico Perazzi and Sergi Caelles and Pablo Arbel\'aez and Alexander Sorkine-Hornung and Luc {Van Gool}},
  title = {The 2017 DAVIS Challenge on Video Object Segmentation},
  journal = {arXiv:1704.00675},
  year = {2017}
}
}

\end{document}